\pdfoutput=1

\documentclass{article}
\usepackage{spconf,amsmath,graphicx,booktabs,url}
\usepackage{cite}

\title{FAIRNESS BEYOND A SINGLE RUN: TRAINING-SEED VARIABILITY\\
IN SPEECH LLM ADAPTATION}

\name{Srishti Ginjala \qquad Eric Fosler-Lussier \qquad Srinivasan Parthasarathy}
\address{Department of Computer Science and Engineering\\
The Ohio State University, Columbus, OH, USA}

\begin{document}
\maketitle

\begin{abstract}
Demographic fairness gaps in automatic speech recognition are almost always reported from a
single training run. We fine-tune the Q-former projector and LoRA adapters of a speech LLM at
five audio compression factors and six random seeds, holding the encoder, base decoder, data
and decoding fixed, and evaluate every run on Common Voice and Fair-Speech. At 460\,h of clean
LibriSpeech, the seed moves fairness metrics more than compression does on most demographic
axes. A balanced $3\times3$ decomposition attributes 85.3\% of the variation in Fair-Speech
ethnicity normalized gap to the seed against 8.3\% to compression ($p=0.009$), though
compression explains more on age and gender. Held-out LibriSpeech word error rate spreads by
0.04 points across those seeds while Common Voice spreads by 8.57, so these are not failed
runs, and the effect survives controlling for accuracy and dropout. Scaling and diversifying
the adaptation set to 960\,h damps the effect but does not remove it. On Fair-Speech ethnicity, two
single-run systems must differ by more than $0.30$ in normalized gap to exceed seed
variability.
\end{abstract}


\section{Introduction}
\label{sec:intro}

Demographic disparities in ASR are well documented, from race-linked gaps in commercial
systems~\cite{koenecke2020racial} to gender and accent gaps in open
models~\cite{tatman2017gender}. Throughout, a \emph{fairness gap} is the spread of word error
rate across demographic groups for a single training run. The reporting convention in this
literature is one trained system per configuration \cite{jahan2025faist,kim2025fairasr,serditova2025newcastle,giraldo2025enhancement,wei-etal-2026-bias,rai2025fairbench,mujtaba2024disfluent,javed2023svarah}. Uncertainty, when reported, is a bootstrap
interval over evaluation utterances: it measures variation across samples, not across
training runs.

In machine learning more broadly, this missing source of uncertainty is known to matter. Ganesh et
al.~\cite{ganesh2023randomness} find that group fairness measures vary sharply across training
runs, with data order the dominant source. D'Amour et al.~\cite{damour2022underspec} call this
underspecification: a training objective can be satisfied equally well by models that behave
differently out of distribution. The effect can be large enough to change a conclusion: in toxic text
classification the seed alone moves a model from the best to the worst of a
fairness-performance tradeoff~\cite{baldini2022fairness}.

Comparable evidence in speech is more limited. Batra et al.~\cite{batra2026seeds} train a
low-resource ASR system
at five seeds and find that several reported gains do not survive replication, but they
measure aggregate word error rate and never break it out by speaker group. Haghbin et
al.~\cite{haghbin2026voice} do put seed error bars on subgroup rates, but for a speech
classification task rather than transcription. Benchmarking work on the corpus we use reports
only evaluation sampling variance and retrains nothing~\cite{herron2026responsible}. A survey
of bias in speech systems does not discuss training randomness at
all~\cite{lin2026survey}. What is missing is a measurement of how much
a per-group word error rate gap moves when the same system is trained again.

Measuring that requires a reference scale, and ours is audio token compression. The choice is not
arbitrary. Group gaps already differ sharply across systems~\cite{koenecke2020racial}, and
Ginjala et al.~\cite{ginjala2026listen} report that compression predicts accent fairness more
strongly than decoder scale. That comparison varied encoder, decoder, adapter
and training data at the same time, so compression was one of several confounded
differences. Here we rebuild that comparison under control: one model, one training
set, one decoding path, and the projector's receptive window as the only thing that moves. The
knob is unusually clean: the projector's parameter shapes are identical at every setting, so
trainable capacity stays fixed while audio tokens per second vary by $2.7\times$. An effect is
expected, since fewer tokens mean coarser acoustic evidence reaching the decoder, which should
cost most on the speech the model already handles worst. We fine-tune the length adapter and LoRA modules of
Granite-Speech-2B~\cite{saon2025granite} across that sweep and repeat the native setting at six
seeds.

To our knowledge this is the first such measurement for ASR. Failed runs, accuracy coupling
and dropout do not explain the spread. It attenuates when the adaptation set is scaled and
diversified, but not on every axis, and the measured spread gives a seed budget for when a
fairness difference is worth interpreting.

\section{Experimental setup}
\label{sec:setup}

\noindent\textbf{Model and design variable.}
Granite-Speech-2B couples a conformer encoder to a
Granite~3.3 2B Instruct decoder through a Q-former projector that maps
\texttt{window\_size} encoder frames to \texttt{num\_queries} audio tokens. The compression
factor is their ratio; the released model uses $15/3=5$. We fine-tune the projector at factors
3, 4, 5, 6 and 8, corresponding to 168, 126, 102, 84 and 63 audio tokens per 10\,s of speech,
a $2.7\times$ span. The projector parameter shapes are identical at every factor, so the
sweep moves the Q-former receptive window while trainable capacity stays fixed.

\smallskip
\noindent\textbf{Warm start.}
Training the projector from random initialization at our data scale collapses to the text
prior, with fluent audio-independent transcripts and dev word error rate above 100\%. The
released projector was trained on roughly 90{,}000\,h. We therefore warm-start every run
from the released factor-5 projector and LoRA. Trainable parameters are 35.7\,M projector plus
17.0\,M LoRA (rank 64, $\alpha=32$, on the decoder \texttt{q\_proj} and \texttt{v\_proj}).
The encoder and the base decoder weights are frozen, but the decoder pathway is adapted, so this is not a
projector-only comparison. Both warm starts are byte-identical across every seed and factor,
so no trainable tensor is randomly initialized and the seed cannot act through
initialization. That scopes the claim to adaptation of a production model, and it should
\emph{reduce} run-to-run variance relative to training from scratch.

\smallskip
\noindent\textbf{Data and seeds.}
The primary rung is LibriSpeech~\cite{panayotov2015librispeech} \texttt{train-clean-100} plus
\texttt{train-clean-360}, 132{,}553 utterances (460\,h). We train for one epoch with batch 8,
gradient accumulation 4, AdamW at $10^{-4}$, a 0.2 warmup ratio and bf16, on one A100 per run,
and take the final checkpoint. Each rung holds 14 distinct runs in two overlapping arms. The
\emph{factor arm} is the full sweep $\{3,4,5,6,8\}$ at seed 1234 and supplies every factor
standard deviation ($n=5$). The \emph{seed grid} crosses factors 3, 5 and 8 with seeds 1234,
2222 and 3333, and factor 5 adds 5678, 1111 and 4444, giving this arm 12 runs; the three runs at seed 1234
belong to both arms. Seed is therefore crossed with factor rather than nested in it, and
factors 4 and 6 exist at seed 1234 alone. The same 14 runs are repeated at 960\,h (adding \texttt{train-other-500}). Two further
rungs, 100\,h and a 460\,h mixture containing \texttt{train-other-500} at matched utterance
count, are used in Section~\ref{sec:ladder} only.

\smallskip
\noindent\textbf{Evaluation.}
We evaluate on Common Voice~24~\cite{ardila2020commonvoice} test (16{,}398 utterances) and
Fair-Speech~\cite{veliche2024fairspeech} (26{,}471 utterances, labelled for ethnicity, gender,
age, first language and socioeconomic status). The Common Voice accent subset has only
2{,}333 labelled utterances with one
group at $n=51$, so we build a balanced expanded accent set of 8{,}998 utterances (1{,}500 per
group, 25 per speaker) from the Common Voice train, dev and test splits, none of
which appear in our adaptation data. Decoding is greedy and identical everywhere.

\smallskip
\noindent\textbf{Metrics.}
With $w_g$ the corpus-level word error rate of group $g$ and $W$ the overall rate,
$\mathrm{MMR}=\max_g w_g/\min_g w_g$ and $\mathrm{NormGap}=(\max_g w_g-\min_g w_g)/W$; we
also report log MMR, worst-group rate and the Gini coefficient over $\{w_g\}$. Ratios average in log
space across seeds, and seed standard deviations are $c_4$-corrected. Bootstrap intervals
measure evaluation sampling variance, and seed spread measures training variance; we never
pool them.

\begin{figure}[t]
\centering
\includegraphics[width=\linewidth]{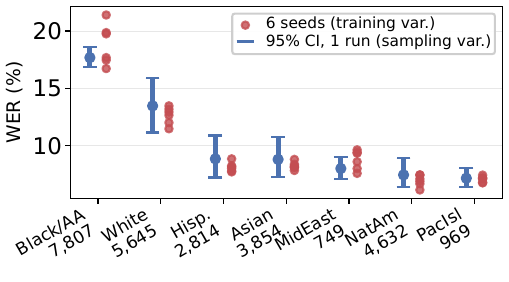}
\caption{Training variance against evaluation sampling variance. Per-group word error rate on
Fair-Speech ethnicity at factor 5, 460\,h. Points are the six
training seeds (training variance); bars are the 95\% bootstrap interval over utterances for
one run (evaluation sampling variance). On the worst-served group the seeds span 4.68 points
against a 1.76 point interval, so training variance is $2.7\times$ wider than the uncertainty
normally reported.}
\label{fig:strip}
\end{figure}

\begin{figure}[t]
\centering
\includegraphics[width=\linewidth]{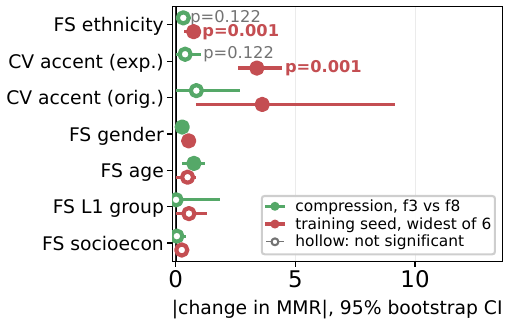}
\caption{Absolute change in MMR from the widest compression pair (factor 3 versus 8) and from
the widest of six seeds, with 95\% bootstrap intervals over utterances, Benjamini-Hochberg
corrected within each family of seven axes. The seed row is an effect-size display and not a
test, for the reason given in Section~\ref{sec:results}. Compression survives on two secondary
axes, in opposite directions.}
\label{fig:bh}
\end{figure}

\section{Results}
\label{sec:results}

\subsection{Seed spread against compression}

Six runs that differ only in seed disagree on how unequally the model treats demographic
groups. Figure~\ref{fig:strip} shows this per group on Fair-Speech ethnicity: on Black/AA
speakers, the worst-served group at every compression factor, training spread is $2.7\times$
wider than the bootstrap interval a single run would report. Conventional fairness tables show
the narrower of the two.

Figure~\ref{fig:bh} places the compression and seed contrasts side by side. After
Benjamini-Hochberg correction~\cite{benjamini1995} within each family of seven axes,
compression survives on two axes (Fair-Speech age, $+0.757$ MMR, $p_{\mathrm{BH}}=0.004$, and gender,
$-0.269$, $p_{\mathrm{BH}}=0.011$). The two effects point in opposite directions, fall on secondary
axes, and do not amount to a mechanism. The seed contrast is the widest pair among six runs,
chosen after seeing the spread, so we read its interval as
an effect size and not as a test; the omnibus evidence for the seed is the permutation test in
Table~\ref{tab:var}, which uses every cell and selects no extreme pair. Across all axis-metric pairs at
460\,h, seed standard deviation exceeds compression standard deviation on 23 of 30 pairs (six axes,
five metrics), excluding overall word error rate.

\subsection{Variance decomposition}

\begin{table}[t]
\centering
\caption{Balanced $3\times3$ factorial at 460\,h: factors $\{3,5,8\}$ crossed with seeds
$\{1234,2222,3333\}$, one run per cell, additive two-way decomposition. One observation per
cell makes the residual the factor\,$\times$\,seed interaction rather than a separate error
term, so no model carrying both is identifiable. The $p$ value comes from permuting seed labels independently within
each factor column 10{,}000 times, breaking the shared seed identity across columns and holding
the factor margin fixed. The last column is the descriptive seed share from the unbalanced
12-run grid.}
\label{tab:var}
\small\setlength{\tabcolsep}{3.2pt}
\begin{tabular}{llrrrr}
\toprule
axis & metric & \% fac. & \% seed & $p$ & 12-run \\
\midrule
FS ethnicity & NormGap & 8.3 & \textbf{85.3} & 0.009 & 83.6 \\
CV accent (exp.) & NormGap & 1.1 & \textbf{91.0} & 0.064 & 91.6 \\
FS ethnicity & MMR & 40.6 & 53.2 & 0.073 & 68.9 \\
FS L1 group & MMR & 15.4 & 59.1 & 0.142 & 63.0 \\
FS socioecon. & MMR & 28.0 & 46.3 & 0.095 & 61.3 \\
FS gender & NormGap & \textbf{87.5} & 9.4 & 0.048 & 52.5 \\
FS age & NormGap & \textbf{90.8} & 4.9 & 0.272 & 17.8 \\
\bottomrule
\end{tabular}
\end{table}

Fair-Speech ethnicity NormGap is the primary axis and metric, fixed before any result was
seen: the rule was to keep it only if its ratio of seed to compression standard deviation
reached $1.5\times$ at 460\,h and held above $1.2\times$ with any single seed dropped, and it
reaches $2.82\times$ and $2.26\times$. Its bootstrap interval
$[1.2, 11.4]$ is wide, six seeds against five factors, but excludes parity at 1. All other
significance claims are secondary.

That ratio compares six training realizations at one design point against five deterministic
design points at one seed; the two are different estimands. Table~\ref{tab:var} therefore takes
the primary inferential role, on the balanced $3\times3$ subset where the two sources are
crossed on equal terms. The primary axis holds there and tightens: seed 85.3\% against
compression 8.3\%, permutation $p=0.009$, the lowest in the table. Compression explains more
than the seed on gender and age, taking 87.5\% and 90.8\%, and we report that rather than
select around it.

\subsection{Alternative explanations for the spread}

\noindent\textbf{In-domain performance.} One seed might simply have trained badly. None did,
and a seed's in-domain performance does not predict its out-of-domain behavior. Across the six
seeds
at 460\,h, held-out LibriSpeech dev-clean word error rate spans 0.04 points while Common Voice
spans 8.57. Normalizing each set's seed standard deviation by its own mean word error rate
gives 1.1\% in-domain easy, 4.7\% in-domain hard (\texttt{test-other}) and 18.6\% on Common
Voice, so difficulty explains part of that rise and domain shift most of it. The seed with
the \emph{best} in-domain word error rate is the worst on Common Voice. A Grubbs test finds no
formal outlier. This is the underspecification signature of D'Amour et
al.~\cite{damour2022underspec}, and it reproduces at $4\times$ the decoder scale
(Granite-Speech-8B, 0.06 in-domain against 2.66 on Common Voice, three seeds).

A second check separates training variance from evaluation noise directly. Expanding the
accent set from 2{,}333 to 8{,}998 utterances narrowed the MMR bootstrap interval by
$9.0\times$ and left the seed standard deviation essentially unchanged (1.273 to 1.219).
Since
utterances cluster by speaker, we also resampled speakers and took all their utterances. The
expanded set draws 8{,}998 utterances from 6{,}048 speakers, so clustering is weak: intervals
move by $0.97\times$ (worst group) and $1.04\times$ (MMR), leaving training spread at
$8.9\times$ the clustered interval. Fair-Speech
provides no speaker identifier, so this check cannot be repeated on the primary axis.

\begin{table}[t]
\centering
\caption{Effect size against accuracy dependence at 460\,h. ``ret.'' is seed variance after
residualizing on overall word error rate over seed variance before; the regression is fitted
over all runs while the variance is taken over the six factor-5 seeds, so the ratio can exceed
1. Thresholds ($1.0\times$, $|r|<0.5$) were fixed before inspection; ``carries'' marks cells
clearing both.}
\label{tab:confound}
\small\setlength{\tabcolsep}{3.4pt}
\begin{tabular}{llrrrl}
\toprule
axis & metric & ratio & $|r|$ & ret. & quadrant \\
\midrule
FS ethnicity & NormGap & 2.82$\times$ & 0.31 & 0.81 & carries \\
FS socioecon. & NormGap & 1.43$\times$ & 0.33 & 1.45 & carries \\
CV accent & MMR & 7.26$\times$ & 0.94 & 0.04 & confound. \\
\bottomrule
\end{tabular}
\end{table}

\smallskip
\noindent\textbf{Gap metrics and overall accuracy.} MMR is a ratio of word error rates, so a seed that
degrades accuracy can inflate it without
treating any group differently. Table~\ref{tab:confound} puts that criticism inside the
result. Expanded accent MMR carries the largest effect in the study at $7.26\times$, and is also the
most confounded: it correlates $0.94$ with overall word error rate and retains 0.04 of its
seed variance after residualizing on accuracy. We therefore do not lead with it. The primary pair sits at
$|r|=0.31$ and retains 0.81 of its variance.

Accuracy dependence is axis-specific: correlation between MMR and overall word error rate
ranges from $0.938$ on expanded accent through $0.514$ on ethnicity to $0.004$ on gender. Any
claim about MMR has to name the axis.

\smallskip
\noindent\textbf{Dropout is not the source of the spread.} Warm start removes initialization as a channel (Section~\ref{sec:setup}), so the seed reaches
the model through only two routes, data order and dropout masks. We retrained four matched
seeds with every dropout probability zeroed and audited the model to confirm dropout was off.
Disabling dropout retains 78\% of the seed standard deviation on the primary axis and 92\% to
96\% on the other three tested, so dropout is not the dominant source. We stop short of
naming data order the sole remaining cause: we did not force deterministic GPU kernels or
dataloader workers, so reduction order is an uncontrolled residual. The result is consistent
with data order dominating, as Ganesh et al.~\cite{ganesh2023randomness} report for
classifiers.

\subsection{Where the seed effect attenuates}
\label{sec:ladder}

Repeating the grid at 960\,h removes the seed's dominance almost everywhere: seed standard
deviation exceeds compression standard deviation on 6 of 30 pairs, against 23 of 30
at 460\,h. Common Voice accent survives, at $1.84\times$ and 70.7\% seed variance
($p=0.032$), as does
Fair-Speech ethnicity NormGap, at $1.08\times$, barely above parity. The primary result is a 460\,h claim.

The attenuation is not a simple volume law. Between 100\,h and 460\,h, both clean-only, seed
$\sigma$ goes \emph{up} by $1.37\times$ ($p=0.026$), which no account based on quantity alone
predicts. A rung separating composition from volume at fixed utterance count points
the same way, but neither leg reaches significance at $n=6$, so we do not rank them.

More epochs do not damp the effect either. At three epochs on the 460\,h rung, seed $\sigma$
rises rather than falls, by $1.20\times$ on the primary axis and $1.68\times$ on expanded
accent over four matched seeds, with in-domain word error rate unchanged. Revisiting each
utterance under three shuffles does not average the effect away, though epochs and total
training length move together here.

\section{A seed budget for fairness reporting}
\label{sec:budget}

\begin{table}[t]
\centering
\caption{Seeds needed per configuration for 80\% power at $\alpha=0.05$, and the smallest
single-run difference exceeding the observed seed spread ($2.77\sigma$), from the six seeds at
factor 5.}
\label{tab:budget}
\small
\begin{tabular}{llrrrr}
\toprule
axis & metric & $\sigma$ & $d{=}0.5$ & $d{=}1.0$ & min.\ diff. \\
\midrule
FS ethnicity & NormGap & 0.108 & 1 & 1 & 0.30 \\
FS ethnicity & MMR & 0.278 & 5 & 2 & 0.77 \\
CV accent & MMR & 1.219 & 94 & 24 & 3.38 \\
FS gender & MMR & 0.193 & 3 & 1 & 0.53 \\
\bottomrule
\end{tabular}
\end{table}

Table~\ref{tab:budget} converts the measured spread into a reporting rule. On the primary
axis at 460\,h two single-run systems must differ by more than $\mathbf{0.30}$ in normalized
gap, or $0.77$ in MMR, before the difference exceeds seed variability, and detecting a true
MMR
difference of 0.5 at 80\% power needs five seeds; at 960\,h the threshold falls to $0.28$ and the requirement to one
seed. All of this rests on a spread estimated from six runs, so the
entries are accurate only to an order of magnitude: the 94 seeds listed for accent at $d{=}0.5$ mean
``more than anyone will train'', not a plan. The requirement moves by more than a factor of two across two rungs
of one ladder, so it has to be estimated per regime rather than borrowed.

The cost of skipping that estimate shows in published work \cite{jahan2025faist,kim2025fairasr,serditova2025newcastle,giraldo2025enhancement,wei-etal-2026-bias,rai2025fairbench,mujtaba2024disfluent,javed2023svarah}, ours
included. In our prior
benchmark~\cite{ginjala2026listen} only 5 of 36 pairwise comparisons on Common Voice accent
(14\%) and 0 of 36 on gender have bootstrap intervals excluding zero, against 47\% to 78\% on
Fair-Speech with roughly $11\times$ the labelled data. Those are evaluation sampling intervals
on released checkpoints, a strictly smaller source of uncertainty than the training variance
measured here.

\section{Limitations and conclusion}
\label{sec:limits}

The study covers one architecture family, English only, adapter training under warm start,
and two main data scales. The 8B replication varies scale within one family and leaves the
second-architecture question open. Freezing the projector and LoRA against each other leaves each arm
retaining at least $0.61$ of the baseline seed spread on the primary axis, so the variance
is not localised to one component. The main grid is a single epoch, with a
three-epoch arm in Section~\ref{sec:ladder}; neither rung is industrial scale. Warm start removes the initialization channel, so our decomposition is narrower than the
from-scratch case. The balanced cells hold
one run each, so the residual in Table~\ref{tab:var} is the interaction, not separable
from error without replicates.

Subject to that scope, the training seed moves fairness metrics further than a
$2.7\times$ compression design variable does on most axes, and growing and diversifying
the adaptation set damps the effect but does not remove it. Conventional validation identifies
performance-unstable seeds, but not fairness-unstable ones. The reporting convention is what we
would change: a per-group word error rate table without a seed floor has left
out its error bars. Per-run metrics and code will be released upon publication.

\vfill\pagebreak
\noindent\textbf{AI use.} AI assistance (Claude) was used for coding and for polishing the
draft. The research idea, the experiments, the evaluation and the write-up are the authors'
own, and the authors take full responsibility for the content of this paper.

\medskip

\bibliographystyle{IEEEbib}
\bibliography{references}

\end{document}